# A Statistical Learning Algorithm for Word Segmentation

Jerry R. Van Aken[*]

ABSTRACT:  In natural speech, the speaker does not pause between words, yet a human listener somehow perceives this continuous stream of phonemes as a series of distinct words. The detection of boundaries between spoken words is an instance of a general capability of the human neocortex to remember and to recognize recurring sequences. This paper describes a computer algorithm that is designed to solve the problem of locating word boundaries in blocks of English text from which the spaces have been removed. This problem avoids the complexities of speech processing but requires similar capabilities for detecting recurring sequences. The algorithm relies entirely on statistical relationships between letters in the input stream to infer the locations of word boundaries. A Viterbi trellis is used to simultaneously evaluate a set of hypothetical segmentations of a block of adjacent words.  This technique improves accuracy but incurs a small latency between the arrival of letters in the input stream and the sending of words to the output stream. The source code for a C++ version of this algorithm is presented in an appendix.

In natural speech, spoken words blend together to form a continuous stream of sounds, yet humans perceive speech as a sequence of distinct words. How the human neocortex accomplishes this feat is an open question.

Even experienced listeners apparently rely, to some extent, on prosodic cues in speech— timing, pauses, stress, and changes in intonation—to separate words and phrases. Is prosody the primary mechanism by which humans learn to identify individual words? Perhaps prosody merely augments a more fundamental learning mechanism that relies on the statistical properties of continuous speech to identify individual words. (See Cutler [4], Kuhl [10], and Saffran, Aslin & Newport [11].) If so, it should be possible to design a computer algorithm to do something similar.

Speech processing introduces complexities that are beyond the scope of this paper. Instead, this paper presents a statistical inference algorithm that addresses a simplified version of this problem: how to detect the word boundaries in a block of English text from which all spaces (and punctuation) are removed. This problem is described as follows.

## *Problem Statement*

Take a block of text, such as the following:

---

[*] Send correspondence to: Jerry Van Aken, Microsoft Corporation, One Microsoft Way, Redmond, WA 98052.

*The quick brown fox jumped over the lazy sleeping dog.*

This example contains ten words, nine of which are unique.

Eliminate all spaces and punctuation from this text, and convert all letters to lower case, as follows:

*thequickbrownfoxjumpedoverthelazysleepingdog*

Design a word-segmentation algorithm that recognizes the individual words in the preceding character stream, and that marks the boundaries between words. The algorithm should produce the following output stream:

*the_quick_brown_fox_jumped_over_the_lazy_sleeping_dog*

A shortcut to achieving such a result is to incorporate a ready-made dictionary that contains some or all of the words that the algorithm will encounter. However, no such shortcut is used here. The algorithm never observes any word in isolation, and it receives no *a priori* information about which combinations of letters are valid words. Instead, the algorithm must discover the words by observing a character stream that is composed of words but that contains no explicit word-boundary information.

A small portion of a character stream that contains words selected at random from the previous example might look like the following:

*...pingjumpedthefoxdogfoxbrownthethedogquickjumpeddoglazyqui...*

This character stream might be generated by a program that randomly selects words from a hidden dictionary and appends them to the stream. The following C++ program generates such a stream:

```
//
// Stochastic stream generator
//
#include <stdlib.h>
#include <stdio.h>

char *test[] =
{
    "the", "quick", "brown", "fox", "jumped",
    "over", "the", "lazy", "sleeping", "dog"
};
const int NUM_TEST_WORDS = sizeof(test)/sizeof(test[0]);
const int NUM_LEARNING_WORDS = 500;

void main()
{
    for (int ix = 0; ix < NUM_LEARNING_WORDS; ++ix)
    {
        int jx = rand() % NUM_TEST_WORDS;
        printf("%s", test[jx]);
    }
}
```

This stochastic stream generator is written in C++ but uses only language features that should be readily understood by a C programmer.

The problem is to design a word-segmentation algorithm that takes, as its input, the character stream that is produced by the stochastic stream generator. This algorithm uses the statistical relationships between the characters in the stream to infer the locations of the word boundaries in the input stream. The algorithm produces, as its output, a modified version of the input stream into which markers are inserted at word boundaries.

The word-segmentation algorithm described in this paper operates in on-line mode. The characters from the input stream appear in the output stream with only a small delay. In contrast, a batch-mode algorithm might not send any characters to the output stream until the algorithm has made one or more passes through the entire input stream.

In the previous C++ program, the state of the stochastic stream generator during each *for*-loop iteration is a combination of a randomly selected word (selected by index $jx$) and the current offset within this word (indicated by index $ix$). The internal state $(ix, jx)$ is hidden. The word-segmentation algorithm must learn to infer the transitions between hidden states by observing only the characters that are generated by these states.

The problem that is addressed by this algorithm is a simplified version of the word boundary problem in continuous speech. The first simplification is to replace the practical problem of finding word boundaries in speech with the somewhat artificial problem of finding word boundaries in English text from which spaces and punctuation are removed. The second simplification is that the words in the input stream to the algorithm occur in random sequence and not according to the rules of grammar. Finally, the probabilities that particular words appear in the character stream are static over the length of the stream. In a real environment, these probabilities might change dynamically over time. An open issue is the extent to which these simplifications might limit the usefulness of this algorithm in real-world applications.

## *Relationship to Prior Work*

Brent [2] provides an overview of word-segmentation algorithms that are primarily published in the speech-processing literature. Gambell & Yang [7] provide a performance comparison many of these algorithms. Goldsmith [8] provides an overview of word-segmentation algorithms that are primarily from the text-processing literature.

Word-segmentation algorithms for speech and text frequently rely on the same statistical regularities to detect words in character streams. However, in speech processing, the focus is primarily on on-line algorithms, whereas the text-processing literature tends to focus more on batch algorithms. The following is a brief description of several representative on-line algorithms from the speech-processing literature.

Elman [5] trained a simple recurrent network (SRN) to predict the next symbol in an input stream based on the symbols that immediately precede this symbol. The input stream consisted of contiguous words selected from a hidden dictionary. The boundaries between words in the input stream were unmarked. The SRN was most successful in predicting the symbols toward the ends of words. The greatest inaccuracy occurred in predicting the first symbol in a new word after reaching the end of the previous word. Thus, increases in the error rate were highly correlated with words, and these increases could be used as hints to locate the word boundaries in the stream.

Saffran, Aslin & Newport [11] studied language learning in infants to determine whether infants can use the statistical relationships between sounds in speech to identify individual words. Saffran et al concluded that infants can segment speech by remembering the transitional probabilities between adjacent phonemes, where the transitional probability that phoneme $y$ immediately follows phoneme $x$ is calculated as the frequency of the sequence $xy$ divided by the frequency of $x$. The transitional probability between two adjacent phonemes tends to be higher if the phonemes are part of the same word, and tends to be lower if the two phonemes are separated by a word boundary.

Brent [2] proposed that the accuracy of word boundary predictions can be improved by using the mutual information between phonemes instead of the simple transitional probabilities that were described by Saffran et al. The mutual information between two phonemes $x$ and $y$ is calculated as $\log_2(P_{xy} / (P_x \cdot P_y))$, where $P_x$ is the frequency of $x$, $P_y$ is the frequency of $y$, and $P_{xy}$ is the frequency of $xy$.

Intuitively, there seems to be a good deal more statistical information in a continuous stream of words than is being exploited by the approaches described in the preceding paragraphs. For example, if one or more familiar words can be successfully identified in a part of the stream, this information might help to reduce uncertainty in identifying other words that share some of the same word boundaries. Additionally, some of the described techniques look at only two adjacent phonemes at a time (although an SRN can, in principle, look more deeply than this into the stream history). The accuracy of the transitional probability of the next symbol in an input stream can be improved by evaluating this probability in the context of a greater number of the symbols that immediately precede it. This principle is used in data compression algorithms such as PPM [3].

Finally, if a word-segmentation algorithm must choose among several competing hypotheses about how to segment some part of an input stream into words, the algorithm's accuracy can be improved by concurrently updating and evaluating all active hypotheses as each new symbol arrives in the input stream. The Viterbi algorithm [6][12] provides a useful means for concurrently evaluating competing hypotheses. Each path in a Viterbi trellis represents a hypothetical sequence of hidden states to account for the input stream observed by the algorithm. The algorithm must limit the number of possible paths that are simultaneously active, and the depth of the history for these paths must be frequently truncated. Otherwise, the storage and processing requirements will quickly become unmanageable for an input stream of any significant length.

The word-segmentation algorithm described in this paper differs from previous algorithms in several ways. The algorithm concurrently evaluates all plausible segmentations of a block of letters that is wide enough to contain several medium-size words. Each such segmentation is a path in a Viterbi trellis. A segmentation path is considered plausible only if it consists entirely of recurring sequences that occur above a certain threshold frequency. A controlled-growth policy limits the number of sequences that the algorithm must store in memory.

Finally, previous algorithms directly calculate next-symbol probabilities based on one or more preceding symbols, and then indirectly infer word boundaries where these probabilities fall beneath some threshold value. In contrast, the algorithm described in this paper uses formulas for transition probabilities that more directly predict the locations of word boundaries.

## *Sequence Memory*

To detect recurring sequences, the word-segmentation algorithm must store sequences that have previously been observed in the input stream and compare these stored sequences to new sequences that arrive in the stream. The algorithm selectively stores sequences in a *sequence memory*, and counts the number of times that each stored sequence is observed.

The sequence memory stores sequences in tree structures. In the following diagram, each node in the tree represents a sequence. The characters in the sequence are contained in this node and in the predecessor nodes. In this example, the sequence *them* is formed by appending one tree node (for the character *m*) to the three tree nodes that form the sequence *the*. The sequence *their* is formed by appending two nodes to the sequence *the*. The leftmost tree node in the diagram is the root of the tree that contains all stored sequences that begin with the character *t*. The sequence memory contains 26 such trees for the letters *a* to *z*. Initially, before the algorithm starts to process the input stream, each tree contains only a root node.

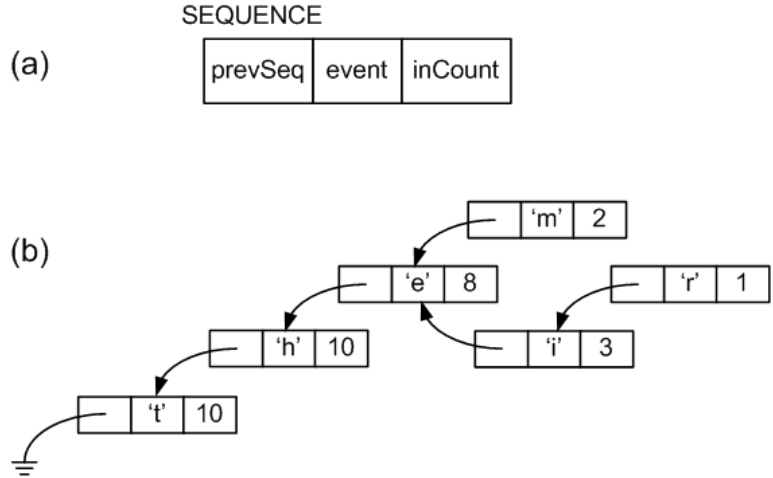

In this diagram, each tree node is a SEQUENCE structure that contains three members. The **event** member contains the letter (character code) at the end of the sequence that is represented by this node. The **inCount** member is a counter that records the number of times this sequence has been observed in the input stream. This member is used to determine the frequency of the sequence. The **prevSeq** member of this structure is a pointer to the node that represents the predecessor sequence. (A SEQUENCE structure has additional members that are described in Appendix A.)

A sequence **X** in the tree has a single predecessor sequence, which is pointed to by **X**.**prevSeq**, but **X** might have numerous successors. The word-segmentation algorithm uses a hash look-up table to find a successor. If $x$ is the character that immediately follows an instance of **X** in the input stream, the algorithm finds the successor **X**′ = **X** + $x$ (where the plus sign indicates concatenation) by calling a **NextSequence** function, which takes **X** and $x$ as arguments and returns **X**′. For example, if the two arguments are the sequence *the* and the character *m*, this function returns the sequence *them*.

### Threshold Frequency

To be practical, the algorithm cannot store every possible sequence that occurs in the input stream. Most are *junk sequences*, which are randomly occurring combinations of parts of adjacent words. A particular junk sequence should repeat only rarely. In contrast, *valid sequences* are actual words or subwords, and these sequences occur relatively frequently. (A subword is a part of a word that is not combined with parts of any other words.)

The algorithm assumes that a sequence is likely to be valid if this sequence is observed to occur above some threshold frequency, $f_T$. The frequency of a sequence **X** in the input stream is calculated as the number of instances of **X** divided by the total number of letters in the stream. In the algorithm described in this paper, $f_T$ is a tuned (set by the programmer) parameter.

For example, the test data for the stochastic stream generator in an earlier section consists of 10 words (including two instances of the word *the*) and 44 letters. The expected frequency of the word *quick* in the generated stream should be about one in 44. However, the junk sequence *ickbro* occurs only when the word *brown* immediately follows the word *quick*, and this junk sequence should have a frequency of about one in 440.

A reasonable value for $f_T$ is $K / N$, where $N$ is the number of letters in the test data, and $K$ is a constant between zero and one. Larger values of $K$ provide for finer discrimination between valid sequences and junk sequences, and might be necessary to achieve accurate word segmentation for large values of $N$ ($> 1000$). However, larger $K$ values slow sequence learning and might necessitate long streams of training data. Smaller values of $K$ enable faster learning of sequences, and work well with small values of $N$ ($< 100$). For the stochastic stream generator discussed in the preceding paragraph ($N = 44$), a value of $K = 0.4$ works well.

The algorithm uses a controlled-growth policy to prevent the sequence memory from filling up with junk sequences. Although most of the sequences in sequence memory are, in fact, junk sequences, every stored sequence consists of a valid sequence to which no more than one additional letter is appended. This restriction prevents the algorithm from having to store longer junk sequences.

If a sequence $\mathbf{X}$ of length $n$ is determined, based on its frequency, to be a valid sequence, the algorithm permits $\mathbf{X}$ to serve as the base sequence for successor sequences of length $n+1$. Therafter, if an instance of $\mathbf{X}$ in the input stream is immediately followed by letter $x$, the new sequence, $\mathbf{X}' = \mathbf{X} + x$, is added to the sequence memory but is tentatively classified as a junk sequence. However, roughly one in ten new sequences are, after a period of monitoring, reclassified as valid sequences, after which they can serve as base sequences for yet longer sequences.

Initially, before the algorithm receives the first letter in the input stream, the sequence memory contains 26 one-letter sequences for the letters *a* to *z*. The algorithm always treats a one-letter sequence as valid, regardless of its frequency.

## *Word Boundary Detection*

To what extent can a learning algorithm that relies on statistical inference alone successfully detect word boundaries in a long sequence of letters that contains no explicit word boundary markers?

Typically, the predictability of the next letter $x_k$ in a sequence $\mathbf{X} = x_1 \ldots x_{k-1}$ can be improved by taking into account one or more of the letters that precede $x_k$ in the sequence. Moreover, a letter at or near the end of a word is easier to predict than a letter at or near the beginning of a word.

In the following diagram, $x_k$ is the last letter in a word of length $N$. As a general rule, a letter $x_k$ at the end of a word of length $N$ is most strongly predicted by the transition probability from the $N$-1 letters that immediately precede it. In contrast, the probability of the transition to $x_k$ from the $N$ or more letters that precede $x_k$ tends to be much weaker because these letters cross a word boundary. Similarly, the probability of the transition to $x_{k+1}$ from the letters that precede $x_{k+1}$ is much weaker because they are separated from $x_{k+1}$ by a word boundary.

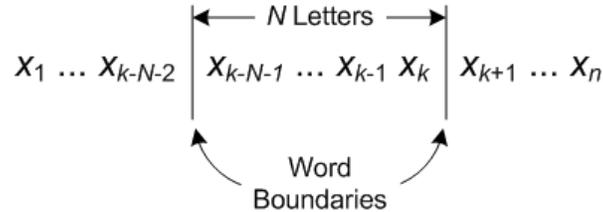

Although the transition probabilities within words can be used with some success to identify individual words in isolation, there are frequent exceptions to the general rule just described. For example, the word *the* occurs with high frequency in English, which makes a word such as *they*, *them* or *then* difficult to discriminate from an instance of the word *the* that is followed by a word that starts with *y*, *m*, or *n*. As a result, a misidentified word boundary might cause a word to be split into two pieces, or parts of two adjacent words might be misidentified as a word.

To avoid such misidentifications, a more reliable strategy is to concurrently identify the word boundaries in a block of several adjacent words. In this way, the probable starting and ending boundaries of a word can be evaluated in the context of the probable boundaries of the adjacent words on either side of this word. This block should start and end on word boundaries so that the first and last words in the block can be reliably identified. The block should be wide enough to contain several words of medium length.

Processing a block of adjacent words is particularly useful for the identification of very short words. A short word is difficult to identify based solely on the transition probabilities within the word. For example, a two-letter word provides only one internal transition probability, and a one-letter word, such as the English word *a*, provides none. However, if a very short word is sandwiched between a pair of relatively long words, and the probable boundaries of these longer words can be determined from their internal transition probabilities, the extent of the short word can then be inferred from these boundaries.

The word-segmentation algorithm uses a special storage structure, called an *event window*, to hold a block of adjacent words. The event window simultaneously evaluates all hypothetical segmentations of a block of contiguous words that can be constructed from the valid sequences that are stored in sequence memory.

As shown in the following diagram, the width of the event window varies over time. When a new letter arrives in the input stream, the algorithm adds a new column to the right side of the event window. This column contains a set of hypothetical word segmentations that include the new letter. When the window contains a sufficient number of columns to enable the algorithm to confidently identify a winning segmentation, one or more words are detached from the left side of the window.

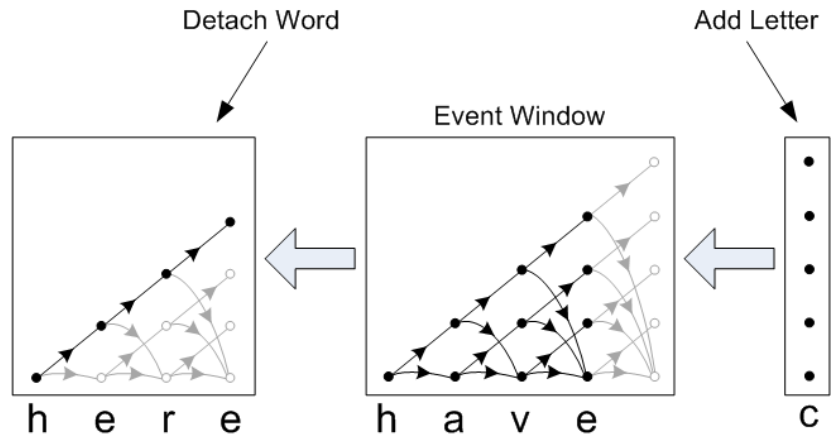

As this diagram indicates, the algorithm is event-driven. The arrival of a new letter in the input stream is an event that triggers a new cycle of processing. This cycle may or may not result in a word being detached from the window and sent to the output stream.

The event window is populated solely with references to valid sequences (words and subwords) that are stored in the sequence memory. A sequence cannot appear in the window until the algorithm has classified the sequence as valid based on the frequency of the sequence.

To identify the word boundaries within a block, the algorithm matches sets of adjacent letters in the block to valid sequences. The algorithm constructs one or more paths through the block, where each path is a set of contiguous sequences that cover the letters in the block, without gaps or overlaps, from one side of the block to the other. Typically, the algorithm can construct a number of such paths through a block from the sequences stored in sequence memory. The sequences in each path represent a hypothetical set of words, and the seams between adjacent sequences in the path are the hypothetical word boundaries.

The example in the following diagram shows the two-dimensional internal structure of the event window. Event time increases from left to right (though probably not in equal time intervals), and sequence length increases from bottom to top. Each solid black dot represents a valid sequence in a hypothetical path. Each column in this diagram contains some number of sequences (hypothetical words) of different lengths. For example, the event window contains three sequences that end with the letter *r*. These three sequences are *r*, *er*, and *her*.

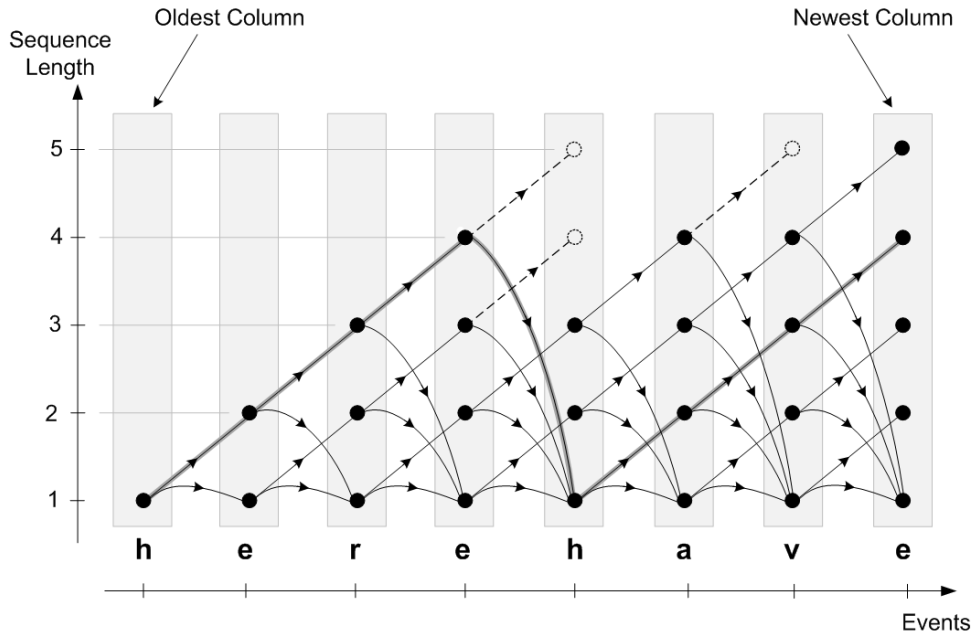

Each solid arrow represents a transition from a valid predecessor sequence to a valid successor sequence. The algorithm assigns a probability to each such transition. A path is a connected set of transitions that spans the width of the window.

A dashed arrow represents a transition to a junk sequence that is stored in sequence memory but that fails the frequency threshold test. The algorithm assigns a probability of zero to such a transition.

In the preceding diagram, the event window contains many hypothetical paths. All paths start at the one-letter sequence, *h*, at the left side of the window. Each path defines a hypothetical set of words. A downward transition in a path represents a hypothetical word boundary. The algorithm assigns a score to each path. The score is the product of the transition probabilities in the path. In this example, the highest-scoring path (highlighted) contains the words *here* and *have*.

The word-segmentation algorithm uses the Viterbi algorithm to calculate the path scores in the event window. As shown in the preceding diagram, the requirements of the word-segmentation algorithm transform of the Viterbi graph into an irregular sawtooth shape instead of the familiar trellis shape.

When a new letter is added to the right edge of the event window, and the new column of the window is populated with valid sequences, the scores for the paths that end in the new column are calculated as the product of the path scores for the previous column and the transition probabilities from the previous column to the new column. The formulas for estimating the transition probabilities are derived in a later section.

In an event window, a sequence **X** of length *n* can have only two possible successor sequences. A successor sequence **X′** of length *n*+1 supports the hypothesis that the letter *x* that immediately follows an instance of **X** in the input stream is a part of the same word as **X**. This is a *same-word* transition. A successor sequence **X′** of length one supports the hypothesis that **X** is a complete word and the letter *x* that immediately follows **X** in the input stream is the start of a new word. This is a *new-word* transition.

A sequence **X** in an event window always has a new-word successor, but may or may not have a same-word successor. The algorithm can always find a valid new-word successor in the sequence memory because all sequences of length one are valid by definition. However, the algorithm adds a same-word successor to the event window only if this successor is found in sequence memory and is a valid sequence.

In the preceding diagram, each sequence of length $n > 1$ always has just one predecessor, which has a length of $n-1$. In contrast, a sequence of length one can have several predecessors, but only the highest-scoring path survives the transition to the start of the new word (by Bellman's principle of optimality), and the algorithm remembers which path is the survivor. Thus, the algorithm needs to assign just one score to each sequence in the event window.

A word is detached from the event window only when the window grows to a minimum width, and the algorithm identifies (based on a *word score*, which is described in a later section) a probable word boundary just to the left of the newest column. A block of contiguous words is defined by this word boundary and by the word boundary at the left edge of the window. Next, the algorithm identifies the probable word boundaries inside this block, detaches the oldest (leftmost) word in the block, and sends this word to the output stream. Thus, the width of the event window shrinks each time a word is detached and sent to the output stream, and grows each time a new letter from the input stream is added to the window. Because the algorithm always detaches a (probable) word from the left side of the window, the left edge of the window is always aligned to a (probable) word boundary.

The word boundary at the right edge of the block cannot be identified as reliably as the boundary at the left edge. The boundary on the left edge was previously defined in the context of the words on either side of this boundary, whereas the context for the boundary on the right edge is still incomplete.

The newer words in the block provide sufficient context to determine the probable word boundaries for the older words in the block. However, after the algorithm detaches one or more of the older words from the block, the remaining words typically have little or no context and cannot be identified with the same high certainty. Before more words can be detached, the algorithm must wait until more letters arrive in the input stream (or the stream ends).

The sequence memory always contains the 26 one-letter sequences $a,b,\ldots,z$. If the algorithm encounters an input sequence that is completely novel, this sequence might, in the extreme case, be sent to the output stream as a series of individual letters, with a word-boundary marker (space) inserted between each pair of letters.

An event window might contain two or more instances of the same sequence. If so, all instances refer to the same stored sequence in sequence memory.

## Transition Probabilities

To calculate the probability of a path through an event window, the algorithm must first calculate the probabilities of the individual sequence-to-sequence transitions in the path. The probability of the full path, from start to end, is the product of the individual transition probabilities in the path.

As discussed previously, if an instance of sequence $\mathbf{X}$ in the input stream is immediately followed by the letter $x$, the two possible transitions from this sequence are the *same-word* transition to sequence $\mathbf{X}' = \mathbf{X} + x$, and the *new-word* transition to the one-word sequence $\mathbf{X}' = 0 + x$, where $\mathbf{X}'$ is the next sequence in the path after $\mathbf{X}$, and 0 represents the null sequence. Thus, only the following two transition probabilities need to be calculated:

- $TP_{same}(\mathbf{X}, x)$, which is the probability of the same-word transition from $\mathbf{X}$ given that $x$ is the value of the letter that immediately follows this instance of $\mathbf{X}$.

- $TP_{new}(\mathbf{X}, x)$, which is the probability of the new-word transition from $\mathbf{X}$ given $x$.

First, $TP_{same}$ can be expressed as follows:

$$
\begin{aligned}
TP_{same}(\mathbf{X}, x) \quad &= \quad P(\mathbf{X}' = \mathbf{X} + x \mid \mathbf{X}) \\
&= \quad P(\mathbf{X}' = \mathbf{X} + x \mid \mathbf{X}' = \mathbf{X} + \square) \cdot P(\mathbf{X}' = \mathbf{X} + \square \mid \mathbf{X})
\end{aligned}
$$

where

- $P(\mathbf{X}' = \mathbf{X} + \square \mid \mathbf{X})$ is the prior probability that whatever letter immediately follows an instance of $\mathbf{X}$ in the input stream (before this letter value is observed) is a part of the same word as $\mathbf{X}$, where the symbol $\square$ is a placeholder for *whatever letter* follows $\mathbf{X}$.

- $P(\mathbf{X}' = \mathbf{X} + x \mid \mathbf{X}' = \mathbf{X} + \square)$ is the conditional probability that $x$ is the value of the letter that immediately follows an instance of $\mathbf{X}$ in the input stream, given that whatever letter follows an instance of $\mathbf{X}$ is a part of the same word as $\mathbf{X}$.

Second, $TP_{new}(\mathbf{X}, x)$ can be expressed as follows:

$$
\begin{aligned}
TP_{new}(\mathbf{X}, x) \quad &= \quad P(\mathbf{X}' = 0 + x \mid \mathbf{X}) \\
&= \quad P(\mathbf{X}' = 0 + x \mid \mathbf{X}' = 0 + \square) \cdot P(\mathbf{X}' = 0 + \square \mid \mathbf{X})
\end{aligned}
$$

where

- $P(\mathbf{X}' = 0 + \square \mid \mathbf{X})$ is the prior probability that whatever letter immediately follows an instance of $\mathbf{X}$ in the input stream is the first letter in a new word, where $\mathbf{X}' = 0 + \square$ is the one-letter sequence that contains this first letter. However, we can immediately relate $P(\mathbf{X}' = 0 + \square \mid \mathbf{X})$ to $P(\mathbf{X}' = \mathbf{X} + \square \mid \mathbf{X})$ as follows:

$$P(\mathbf{X}' = 0 + \square \mid \mathbf{X}) \; = \; 1.0 - P(\mathbf{X}' = \mathbf{X} + \square \mid \mathbf{X})$$

- $P(\mathbf{X}' = 0 + x \mid \mathbf{X}' = 0 + \square)$ is the conditional probability that $x$ is the value of the first letter in a new word, given that whatever letter follows $\mathbf{X}$ is the start of a new word.

Thus, to calculate the transition probabilities $TP_{same}$ and $TP_{new}$, the algorithm must first calculate the prior probability $P(\mathbf{X}' = \mathbf{X} + \square \mid \mathbf{X})$, and the conditional probabilities $P(\mathbf{X}' = \mathbf{X} + x \mid \mathbf{X}' = \mathbf{X} + \square)$ and $P(\mathbf{X}' = 0 + x \mid \mathbf{X}' = 0 + \square)$. The word-segmentation algorithm uses approximations for these three probabilities. The accuracy of the algorithm largely depends on the accuracy of these approximations.

To approximate the prior probability $P(\mathbf{X}' = \mathbf{X} + \square \mid \mathbf{X})$, the algorithm must gather some additional statistics for each sequence. For this purpose, two new counters are added to the previously described SEQUENCE structure. In addition to the **inCount** and **createCount** members, which were previously discussed, this structure is augmented with two new members, **outCount** and **succCount**.

For a stored sequence $\mathbf{X}$, the **outCount** member counts the number of valid instances of $\mathbf{X}$. Each time the algorithm observes an instance of $\mathbf{X}$ in the input stream, and the calculated frequency of $\mathbf{X}$ at the time of this observation exceeds the threshold frequency $f_T$, $\mathbf{X}$.**outCount** is incremented by one. If the length of $\mathbf{X}$ is greater than one, the **succCount** member of the predecessor of $\mathbf{X}$ is incremented by one at the same time $\mathbf{X}$.**outCount** is incremented.

The **succCount** member of sequence $\mathbf{X}$ counts the number of times that a same-word successor to $\mathbf{X}$ is a valid sequence. As discussed previously, a valid sequence in sequence memory can have several successors (for example, the word *the* can have successors *them*, *they*, and so on). If the $k$ successors to $\mathbf{X}$ are $\mathbf{X}'_i = \mathbf{X} + x_i$, $i = 1,\ldots,k$, then $\mathbf{X}$.**succCount** is equal to the following sum:

$$\mathbf{X}.\textbf{succCount} \; = \; \sum_{i=1}^{k} \mathbf{X}'_i.\textbf{outCount}$$

If $x$ is the letter value that immediately follows an instance of sequence $\mathbf{X}$, and the algorithm determines that the successor $\mathbf{X}' = \mathbf{X} + x$ is a valid sequence, both $\mathbf{X}$.**succCount** and $\mathbf{X}'$.**outCount** are incremented by one, and this occurrence indicates that $\mathbf{X}$ and $x$ are probably parts of the same word. If $\mathbf{X}$.**outCount** is incremented, but $\mathbf{X}$.**succCount** and $\mathbf{X}'$.**outCount** are not incremented, this occurrence probably indicates that the letter $x$ that follows $\mathbf{X}$ is the start of a new word, and that a word boundary separates $\mathbf{X}$ from $\mathbf{X}'$.

Conceptually, incrementing $\mathbf{X}$.**outCount** is analogous to the firing of a neuron that represents sequence $\mathbf{X}$. Each time $\mathbf{X}$.**inCount** is incremented by an input event, the algorithm must determine, based on the frequency of $\mathbf{X}$, whether $\mathbf{X}$ will fire. If $\mathbf{X}$ fires, the algorithm must then determine whether the successor sequence $\mathbf{X}'$ will fire, and so on.

After **X** fires, a successor **X′** = **X** + $x$ will fire only if the following conditions are satisfied:

- The letter that immediately follows this instance of **X** in the input stream is $x$. (This condition always causes **X′.inCount** to increment regardless of whether **X′.outCount** increments.)

- The frequency of **X′** exceeds threshold frequency $f_T$.

However, the one-letter sequence **X′** = 0 + $x$, which has no predecessor, fires unconditionally each time an instance of $x$ is observed in the input stream.

Ideally, **X′.outCount** counts the number of times that successor **X′** = **X** + $x$ is part of the same word as **X**, and **X.succCount** counts the number of times *any* successor to **X** is part of the same word as **X**. However, both of these counters are prone to overcounting. The reason is that the algorithm determines whether a successor sequence is valid solely based on its frequency. Thus, if **X′** = **X** + $x$ is recognized as a valid sequence but, for the current instance of **X** and $x$, the letter $x$ is, in fact, the start of a new word, **X′.outCount** and **X.succCount** are still incremented.

For example, if the input stream contains instances of the words *the*, *them*, and one or more words that begin with the letter *m*, then, immediately after *m* is observed to follow an instance of *the*, there is uncertainty about whether this *m* is part of the same word as *the*, or is the start of a new word. In either case, the algorithm increments **X.succCount** for **X** = *the* and **X′.outCount** for **X′** = *them* because the algorithm classifies **X′** as a valid sequence based solely on its frequency.

In the following discussion, δ represents the overcount amount in **X.succCount**, and ε represents the overcount amount in **X′.outCount**. The following is the prior probability that **X** and whatever letter follows **X** are part of the same word:

$$P(\mathbf{X'} = \mathbf{X} + \square \mid \mathbf{X}) = \frac{\mathbf{X.succCount} - \delta}{\mathbf{X.outCount}}$$

The following is the conditional probability that $x$ is the value of the letter that immediately follows an instance of sequence **X**, given that whatever letter follows an instance of **X** is part of the same word as **X**:

$$P(\mathbf{X'} = \mathbf{X} + x \mid \mathbf{X'} = \mathbf{X} + \square) = \frac{\mathbf{X'.outCount} - \varepsilon}{\mathbf{X.succCount}}$$

Finally, if the input stream, thus far, contains a total of $N$ letters, and $n_x$ of these letters are letter $x$, the following approximation is used for the probability that a word starts with $x$:

$$P(\mathbf{X'} = 0 + x \mid \mathbf{X'} = 0 + \square) = \frac{n_x}{N}$$

This approximation is based on the assumption that the probability that a letter occurs at the start of a word is the same as the probability that a letter occurs anywhere in the input stream. The validity of this assumption varies according to the characteristics of the hidden dictionary that is used to generate the input stream to the algorithm.

Estimates of the overcount amounts $\delta$ and $\varepsilon$ are required to calculate the probabilities $P(\mathbf{X}' = \mathbf{X} + \square \mid \mathbf{X})$ and $P(\mathbf{X}' = \mathbf{X} + x \mid \mathbf{X}' = \mathbf{X} + \square)$. The following difference is the number of times that whatever letter follows $\mathbf{x}$ is the start of a new word:

$$diff = \mathbf{X}.\mathbf{outCount} - (\mathbf{X}.\mathbf{succCount} - \delta)$$

The ratio $\varepsilon \, / \, diff$ must be very close to the ratio $n_x / N$, where, as before, $n_x$ is the number of instances of letter $x$ in the input stream and $N$ is the total number of letters in the stream. If we assume that $\delta << diff$, the value of $\varepsilon$ can be estimated as follows:

$$\varepsilon \approx (\mathbf{X}.\mathbf{outCount} - \mathbf{X}.\mathbf{succCount}) \cdot \frac{n_x}{N}$$

The value of $\delta$ is more difficult to estimate, although it must be at least as large as $\varepsilon$. However, the ratio of $\delta$ to $\mathbf{X}.\mathbf{succCount}$ tends to be much smaller than the ratio of $\varepsilon$ to $\mathbf{X}'.\mathbf{outCount}$, so $\delta$ has significantly less impact than $\varepsilon$ on the transition probabilities. Based on this rationale, the algorithm sets $\delta = 0$.

By substituting these values for $\delta$ and $\varepsilon$ into the previous formulas for $P(\mathbf{X}' = \mathbf{X} + \square \mid \mathbf{X})$ and $P(\mathbf{X}' = \mathbf{X} + x \mid \mathbf{X}' = \mathbf{X} + \square)$, the following approximate formulas are derived for the transition probabilities:

$$P_{new}(\mathbf{X}, x) \approx \left(1.0 - \frac{\mathbf{X}.\mathbf{succCount}}{\mathbf{X}.\mathbf{outCount}}\right) \cdot \frac{n_x}{N}$$

$$P_{same}(\mathbf{X}, x) \approx \frac{\mathbf{X}'.\mathbf{outCount}}{\mathbf{X}.\mathbf{outCount}} - P_{new}(\mathbf{X}, x)$$

These formulas are used in the program listing for the word-segmentation algorithm in Appendix B. Occasionally, the value of $P_{same}$ is slightly less than zero because of the subtraction of $P_{new}$, but this error seems to have little or no effect on performance, and the algorithm does not bother to clamp negative values of $P_{same}$ to zero.

## Two-Level Scoring System

The transition probabilities that were derived in the previous section are used to calculate the first-level path scores in what is a two-level scoring system. As discussed previously, the (first-level) score for a path is calculated as the product of the transition probabilities in the path. These first-level scores, by themselves, are somewhat context-sensitive, and segmentations that are based solely on these scores are subject to occasional errors.

A second scoring level is introduced to produce more consistent and reliable results. The second-level scores are essentially a refined version of the first-level scores.

Although the first-level scores are usually reliable, they are sometimes degraded by word combinations that create ambiguities, and these ambiguities can cause the algorithm to incorrectly identify word boundaries. In particular, the letters at the end of a word in a block might occasionally combine with the letters at the start of the next word in the block to match a stored sequence, and thereby fool the algorithm into misidentifying the word boundaries in the block.

The following diagram indicates how the first-level scores are calculated when a new letter arrives in the input stream and a column is added to the right side of the event window.

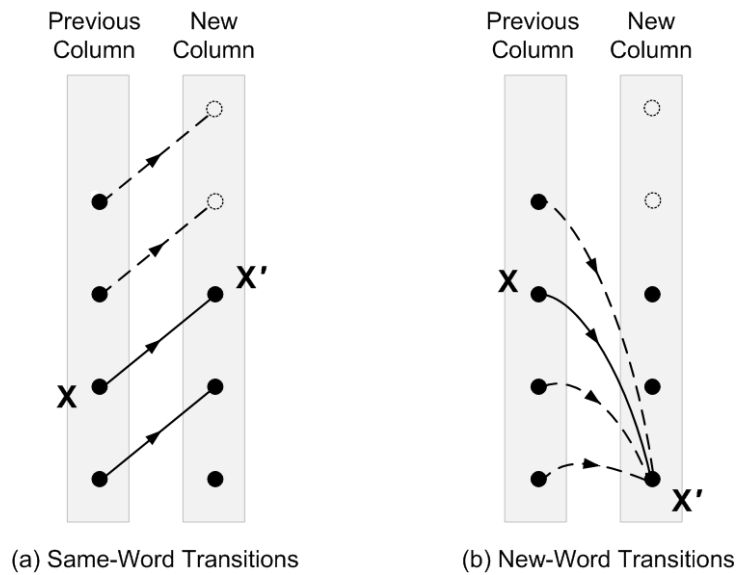

Part (a) of this diagram shows the same-word transitions from the previous column in the event window to the new column. Dashed arrows represent transitions to junk sequences, and these transitions are assigned a probability of zero (and discarded). Solid arrows represent transitions to valid sequences, and the probability of each such transition is calculated by using the formula for $TP_{same}$ that was derived in the previous section. The raw score for the path that ends at $\mathbf{X}'$ and that includes the transition from sequence $\mathbf{X}$ in the previous column is the product of the transition probability and the score that was previously calculated for the path that ends at $\mathbf{X}$.

Part (b) of the preceding diagram shows the new-word transitions. Several transitions are shown from sequences in the previous column to the one-letter sequence $\mathbf{X}'$ in the new column, but only the highest-scoring path survives this transition. In this case, dashed arrows represent transitions in the discarded paths, and the solid arrow represents the transition from $\mathbf{X}$ to $\mathbf{X}'$ in the survivor path. The probability of the transition from $\mathbf{X}$ to $\mathbf{X}'$ is calculated by using the formula for $TP_{new}$ that was derived in the previous section. The raw score for the path that includes this transition is the product of the transition probability and the score that was previously calculated for the path that ends at **X**.

After the raw first-level scores are calculated for all of the sequences in the new column, as described in the two preceding paragraphs, these scores are normalized so that the sum of the scores in the column is one. After normalization, these first-level scores are ready to propagate to the second level of the two-level scoring system.

In part (b) of the previous diagram, the survivor path that ends at the one-letter sequence **X′** in the new column includes the new-word transition from **X** in the previous column. The algorithm uses the normalized first-level score for this path as the estimated probability that this instance of **X** is a complete word.

Each time an instance of **X** is part of a survivor path to a one-letter sequence, the first-level score for this path is added to the accumulated word score for **X**, which is stored in the **accumScores** member of the SEQUENCE structure for **X**. After $n_X$ instances of a sequence **X** have been observed in the input stream, the *average word score* for **X** is calculated as **X**.**accumScores** / $n_X$. This average word score is the estimated probability that an instance of **X** is a word. Because the first-level scores are normalized before they are added to **X**.**accumScores**, the average word score is a value between zero and one.

The second-level score for a path is calculated from the average word scores of all of the hypothetical words in the path. To calculate the second-level score for a path, the algorithm first calculates the partial score for each hypothetical word **X** in the path as the product of the average word score of **X** and the length of **X**. The second-level score for the path is then the product of these partial scores.

## Discussion

The program listing in Appendix B contains a complete implementation of the word-segmentation algorithm, plus a test program to exercise the algorithm. The test data for this program is Lincoln's Gettysburg Address. This speech is 271 words in length and contains a total of 1,149 characters. Some words appear multiple times. For example, the word *here* is used eight times. The number of unique words in the speech is 138.

The output stream that the algorithm generates for this test data contains several errors. The algorithm fails to insert word boundary markers (spaces) between two instances of the words *in_a* and one instance of *on_a*. These errors are symptoms of an inherent weakness in the algorithm. To identify a very short word like *in*, *on* or *a*, the algorithm relies on the longer words that surround the short word to help to establish the word boundaries around the short word. However, when two very short words are adjacent to each other, this strategy sometimes fails.

As explained in an earlier section, the threshold frequency $f_T$ is set to $K / N$, where $N$ is the number of words in the test data, and $K$ is a tuned constant between 0.5 and 1.0. For the Gettysburg Address example, $N$ is 271 and $K$ is 0.76.

The $K$ value for this example is highly tuned. For example, increasing $K$ from 0.76 to 0.765 introduces an additional error. Namely, the word *highly* is incorrectly segmented as *high_ly*. The problem here is that this word is the only one in the speech that begins with *hi*. In addition, the speech contains only one instance of the word *highly*. Because of the algorithm's policy of controlled growth, the representation of this word in the sequence memory grows very slowly and is still incomplete after the algorithm has processed an input stream of 175,000 words. This error can be eliminated by increasing the value of NUM_LEARNING_WORDS from 175,000 to 250,000.

The test program randomly selects words to append to the input stream to the algorithm. For a test case, such as the Gettysburg Address, that contains over a thousand letters, the error rate of the algorithm is sensitive to the characteristics of the random number generator. The specific errors discussed in the preceding paragraphs are based on the seed value in the program listing, and on the **srand** and **rand** functions in the Microsoft C/C++ runtime library.

A good random number generator is just as likely to pick one word as another from the test data. Thus, if the test data contains $N$ letters and just one instance of the word *abcd*, the nominal frequency of *abcd* is $1 / N$. However, this word might appear in the input stream at a frequency that is somewhat higher or lower than the nominal frequency. In fact, the distribution of word frequencies that results from the use of the random number generator takes the form of a bell-shaped curve that is centered at the nominal frequency, $1 / N$. The few words that lie in the extreme low-frequency end of this bell curve might occur at a frequency that is significantly lower than the nominal frequency.

In the Gettysburg Address example, the error rate of a word, such as the word *here*, that appears several times in the speech is largely immune to placement in the low-frequency end of the bell-shaped curve. On the other hand, the error rate of a word, such as *highly*, that appears only once in the speech is highly sensitive to such placement.

In Appendix B, the **SetFirstLevelScores** function calculates the first-level scores for a new column, and the **SetSecondLevelScores** function calculates the second-level scores for an entire event window. As previously explained, the algorithm selects word boundaries based on the second-level scores. The accumulated first-level scores are used as intermediate values in the calculation of the second-level scores. However, the algorithm is easily modified to use just the first-level scores to select the word boundaries—simply comment out the call to **SetSecondLevelScores**. With this change, the accuracy of the algorithm declines noticeably.

The **PopulateNewColumn** function populates a new column in the event window with pointers to valid sequences. In addition, this function updates the statistical counts for the sequences in sequence memory, and adds new sequences to this memory.

**PopulateNewColumn** is responsible for nearly the entire construction and maintenance of the sequence memory. The only exception is that the **accumScores** values in sequence memory are updated by the **SetFirstLevelScores** function so that these values are available for use by the **SetSecondLevelScores** function. Thus, virtually all of the learning done by the algorithm occurs in **PopulateNewColumn**.

For the Gettysburg Address example, the word-segmentation algorithm stores 14,543 sequences in sequence memory. Of these, only 1,532 are classified as valid sequences. Some applications might require the algorithm's storage requirements to be reduced. A strategy for reclaiming sequence memory is to delete a stored junk sequence after it has been monitored for long enough to confirm that it cannot be a valid sequence. The storage for this sequence can then be recycled. Another storage-reduction strategy is to use a smaller version of the SEQUENCE structure for newly created sequences. A stored sequence that is later confirmed to be valid can be upgraded to use a larger SEQUENCE structure that has a full set of counters.

## Appendix A: The SEQUENCE Structure

The SEQUENCE structure represents a sequence of characters. Each sequence that is stored in sequence memory is represented by a SEQUENCE structure. This structure is defined as follows:

```
typedef struct _SEQUENCE
{
    struct _SEQUENCE *_link;
    struct _SEQUENCE *_prevSeq;
    struct
    {
        BYTE _event;
        BYTE _length;
        WORD _allocNum;
    } _info;
    DWORD _createCount;
    DWORD _outCount;
    DWORD _inCount;
    DWORD _succCount;
    float _accumScores;
} SEQUENCE;
```

## Members

**link**

A pointer to the next SEQUENCE structure in a linked list. A hash table contains all of the sequences in sequence memory. Each bucket in the hash table contains a linked list of sequences.

**prevSeq**

    A pointer to the SEQUENCE structure that represents the predecessor sequence in sequence memory.

**event**

    Specifies the ANSI character code at the end of the sequence that is represented by this SEQUENCE structure. The other characters in this sequence are contained in the predecessor sequence that is pointed to by the **prevSeq** member.

**length**

    Specifies the number of characters in the sequence that is represented by this SEQUENCE structure.

**allocNum**

    Specifies the allocation number that is assigned to this structure. Each created instance of the SEQUENCE structure is assigned an allocation number to uniquely identify the instance.

**createCount**

    Specifies the number of characters that had been received in the input stream at the time that this SEQUENCE structure instance was created. The frequency at which instances of the sequence represented by this structure are observed in the input stream is measured relative to the **createCount** value.

**inCount**

    Specifies the number of instances of the sequence represented by this structure that have been observed in the input stream.

**outCount**

    Specifies the number of observed instances of the sequence represented by this structure that have been determined to be valid sequences, based on the frequency at which the sequence appears in the input stream. This structure's **outCount** value is always less than its **inCount** value.

**succCount**

    Specifies the number of times that the sequence represented by this structure is immediately followed by a character that is part of the same word as this sequence. This structure's **succCount** value is incremented by one each time an immediate successor's **outCount** value is incremented by one. Thus, this structure's **succCount** value equals the sum of the **outCount** values of all of its immediate successors. This structure's **succCount** value is always less than its **outCount** value.

**accumScores**

    Specifies the accumulated first-level scores for the sequence that is represented by this structure. Each time an instance of this sequence is the most likely sequence

in an event window column to be a complete word, the first-level score for this sequence is added to **accumScores**. (The **accumScores** members of the other sequences in the column remain unchanged.) The average value of these accumulated scores, which is obtained by dividing this structure's **accumScores** value by its **inCount** value, indicates the relative probability that this sequence is a word.

## *Appendix B: Program Listing*

The following source code listing contains an implementation of the word-segmentation algorithm, and includes a test program and test data. This listing is a complete program that is written in C++ but uses only language features that should be readily understood by a C programmer. To run this program, first copy all of the source code into a file that has a .cpp file name extension. Then compile the program and run it.

```
//
// Word-segmentation algorithm
//

#include <stdlib.h>
#include <string.h>
#include <memory.h>
#include <stdio.h>
#include <assert.h>
#include <math.h>

#define ARRAY_LENGTH(x)   (sizeof(x)/sizeof(x[0]))

const int MAX_WINDOW_WIDTH = 32;

typedef unsigned char UINT8;
typedef unsigned short UINT16;
typedef unsigned long UINT32;

// A sequence stored in sequence memory
typedef struct _SEQUENCE
{
    struct _SEQUENCE *_link;
    struct _SEQUENCE *_prevSeq;
    struct
    {
        UINT8 _event;
        UINT8 _length;
        UINT16 _allocNum;
    } _info;
    UINT32 _createCount;
    UINT32 _inCount;
    UINT32 _outCount;
    UINT32 _succCount;
    float _accumScores;
} SEQUENCE;

// A cell in a column in an event window
typedef struct _CELL
```

```c
{
    SEQUENCE *_seq;
    float _score;
} CELL;

// A column in an event window
typedef struct _COLUMN
{
    struct _COLUMN *_prevCol;
    struct _COLUMN *_nextCol;
    UINT8 _event;
    UINT8 _bestLength;
    float _bestScore;
    int _numCells;
    CELL _cell[MAX_WINDOW_WIDTH];
} COLUMN;

// Global variables used by word-segmentation algorithm
const float THRESHOLD_BIAS = 4.567;
SEQUENCE _rootSequence['z' - 'a' + 1];
SEQUENCE *_hashTable[12577];
COLUMN _column[MAX_WINDOW_WIDTH];
COLUMN *_headColumn = NULL;
UINT16 _allocCount = 0;
int _eventCount = 0;
int _fireCount = 0;
int _minColumns = MAX_WINDOW_WIDTH / 2;
int _numColumns = 0;
int _enableWrite = 0;

//
// Test data and tuned parameters
//
#if 0

const int NUM_LEARNING_WORDS = 500;
const float THRESHOLD_PROB = 0.4 / 44.0;
char *_testData[] =
{
    "the", "quick", "brown", "fox", "jumped",
    "over", "the", "lazy", "sleeping", "dog"
};

#else

const int NUM_LEARNING_WORDS = 175000;
const float THRESHOLD_PROB = 0.76 / 1149.0;
char *_testData[] =
{
    "four", "score", "and", "seven", "years", "ago", "our", "fathers",
    "brought", "forth", "on", "this", "continent", "a", "new",
    "nation", "conceived", "in", "liberty", "and", "dedicated", "to",
    "the", "proposition", "that", "all", "men", "are", "created",
    "equal", "now", "we", "are", "engaged", "in", "a", "great",
    "civil", "war", "testing", "whether", "that", "nation", "or",
    "any", "nation", "so", "conceived", "and", "so", "dedicated",
    "can", "long", "endure", "we", "are", "met", "on", "a", "great",
    "battlefield", "of", "that", "war", "we", "have", "come", "to",
    "dedicate", "a", "portion", "of", "that", "field", "as", "a",
    "final", "resting", "place", "for", "those", "who", "here",
    "gave", "their", "lives", "that", "that", "nation", "might",
    "live", "it", "is", "altogether", "fitting", "and", "proper",
    "that", "we", "should", "do", "this", "but", "in", "a", "larger",
```

```
        "sense", "we", "can", "not", "dedicate", "we", "can", "not",
        "consecrate", "we", "can", "not", "hallow", "this", "ground",
        "the", "brave", "men", "living", "and", "dead", "who",
        "struggled", "here", "have", "consecrated", "it", "far", "above",
        "our", "poor", "power", "to", "add", "or", "detract", "the",
        "world", "will", "little", "note", "nor", "long", "remember",
        "what", "we", "say", "here", "but", "it", "can", "never",
        "forget", "what", "they", "did", "here", "it", "is", "for", "us",
        "the", "living", "rather", "to", "be", "dedicated", "here", "to",
        "the", "unfinished", "work", "which", "they", "who", "fought",
        "here", "have", "thus", "far", "so", "nobly", "advanced", "it",
        "is", "rather", "for", "us", "to", "be", "here", "dedicated",
        "to", "the", "great", "task", "remaining", "before", "us", "that",
        "from", "these", "honored", "dead", "we", "take", "increased",
        "devotion", "to", "that", "cause", "for", "which", "they", "gave",
        "the", "last", "full", "measure", "of", "devotion", "that", "we",
        "here", "highly", "resolve", "that", "these", "dead", "shall",
        "not", "have", "died", "in", "vain", "that", "this", "nation",
        "under", "god", "shall", "have", "a", "new", "birth", "of",
        "freedom", "and", "that", "government", "of", "the", "people",
        "by", "the", "people", "for", "the", "people", "shall", "not",
        "perish", "from", "the", "earth"
};

#endif

//
// Initialize data structures used by the word-segmentation algorithm.
//
void Initialize()
{
    int ix;

    memset(_hashTable, 0, sizeof(_hashTable));
    memset(_rootSequence, 0, sizeof(_rootSequence));
    for (ix = 0; ix < ARRAY_LENGTH(_rootSequence); ++ix)
    {
        _rootSequence[ix]._info._event = ix + 'a';
        _rootSequence[ix]._info._length = 1;
        _rootSequence[ix]._info._allocNum = ++_allocCount;
        _rootSequence[ix]._accumScores = 0.0;
    }

    memset(_column, 0, sizeof(_column));
    for (ix = 1; ix < MAX_WINDOW_WIDTH; ++ix)
    {
        _column[ix-1]._nextCol = &_column[ix];
        _column[ix]._prevCol = &_column[ix-1];
    }
    _column[MAX_WINDOW_WIDTH - 1]._nextCol = &_column[0];
    _column[0]._prevCol = &_column[MAX_WINDOW_WIDTH - 1];
}

//
// Use the ELF hash algorithm to generate a hash table index.
//
UINT32 GetHashIndex(SEQUENCE *seq, UINT8 event)
{
    const UINT32 count = sizeof(event) + sizeof(seq->_info);
    UINT8 *p = (UINT8*)(&seq->_info);
    UINT32 h = 0, g;

    for (int ix = 0; ix < count; ++ix)
```

```c
    {
        h = (h << 4) + event;
        if (g = h & 0xF0000000)
            h ^= g >> 24;

        h &= ~g;
        event = *p++;
    }
    return (h % ARRAY_LENGTH(_hashTable));
}

//
// Look up the sequence that is formed by appending a character
// to the specified input sequence.
//
SEQUENCE* NextSequence(SEQUENCE *seq, UINT8 event)
{
    SEQUENCE *p, *q = NULL;
    int index;

    assert('a' <= event && event <= 'z');
    if (!seq)
        return &_rootSequence[event - 'a'];

    index = GetHashIndex(seq, event);
    p = _hashTable[index];
    if (p)
    {
        if (p->_prevSeq == seq && p->_info._event == event)
            return p;

        for (q = p, p = p->_link; p; q = p, p = p->_link)
            if (p->_prevSeq == seq && p->_info._event == event)
                break;

        if (p)
            q->_link = p->_link;
    }
    if (!p)
    {
        p = (SEQUENCE*)malloc(sizeof(SEQUENCE));
        assert(p);
        memset(p, 0, sizeof(SEQUENCE));
        p->_prevSeq = seq;
        p->_info._event = event;
        p->_info._length = 1 + seq->_info._length;
        p->_info._allocNum = ++_allocCount;
        p->_createCount = _eventCount;
    }
    p->_link = _hashTable[index];
    _hashTable[index] = p;
    return p;
}

//
// Populate a new column in the event window with pointers to valid
// sequences that match the latest characters to arrive in the input
// stream. Additionally, update the statistics in the sequence memory.
//
int PopulateNewColumn(UINT8 event, CELL inCell[],
                      int numInCells, CELL outCell[])
{
    int ix, numOutCells;
```

```
    SEQUENCE *inSeq, *outSeq;

    ++_eventCount;
    outSeq = NextSequence(NULL, event);
    outSeq->_inCount += 1;
    outSeq->_outCount += 1;
    outCell[0]._seq = outSeq;
    outCell[0]._score = 0.0;
    numOutCells = 1;
    for (ix = 0; ix < numInCells; ++ix)
    {
        float prob;

        inSeq = inCell[ix]._seq;
        outSeq = NextSequence(inSeq, event);
        outSeq->_inCount += 1;
        prob = (float)(outSeq->_inCount - THRESHOLD_BIAS) /
                (_eventCount - outSeq->_createCount);
        if (prob < THRESHOLD_PROB)
            break;

        if (!outSeq->_outCount)
            ++_fireCount;

        outSeq->_outCount += 1;
        inSeq->_succCount += 1;
        outCell[numOutCells]._seq = outSeq;
        ++numOutCells;
        assert(numOutCells < MAX_WINDOW_WIDTH);
        assert(outSeq->_info._length == numOutCells);
    }

    // Update the statistics on these probable junk sequences,
    // because a few of them will turn out to be valid sequences.
    for (++ix ; ix < numInCells; ++ix)
    {
        inSeq = inCell[ix]._seq;
        outSeq = NextSequence(inSeq, event);
        outSeq->_inCount += 1;
    }
    return numOutCells;
}

//
// Calculate first-level scores for the valid sequences in the new column.
//
void SetFirstLevelScores()
{
    COLUMN *col = _headColumn->_prevCol;
    CELL *inCell = col->_cell;
    CELL *outCell = _headColumn->_cell;
    float frac = (float)outCell[0]._seq->_inCount / _eventCount;
    float sum = 0.0;

    assert(_headColumn->_numCells > 0);
    for (int ix = 0; ix < col->_numCells; ++ix)
    {
        SEQUENCE *inSeq = inCell[ix]._seq;
        float score = inCell[ix]._score;
        float eowCount = inSeq->_outCount - inSeq->_succCount;
        float Pnew = frac * eowCount / inSeq->_outCount;

        if (ix + 1 < _headColumn->_numCells)
```

```
        {
            SEQUENCE *outSeq = outCell[ix+1]._seq;
            float Psame = (float)outSeq->_outCount / inSeq->_outCount - Pnew;

            outCell[ix+1]._score = score * Psame;
            sum += outCell[ix+1]._score;
        }
        score *= Pnew;
        if (!ix || outCell[0]._score < score)
        {
            outCell[0]._score = score;
            col->_bestLength = ix + 1;
        }
    }

    // Normalize first-level scores in new column.
    if (sum != 0.0)
    {
        sum += outCell[0]._score;
        for (int jx = 0; jx < _headColumn->_numCells; ++jx)
            outCell[jx]._score /= sum;
    }
    else
        outCell[0]._score = 1.0;

    // Remember survivor path to start of hypothetical new word.
    inCell[col->_bestLength-1]._seq->_accumScores += outCell[0]._score;
}

//
// Calculate second-level scores for the paths in the event window.
//
void SetSecondLevelScores()
{
    COLUMN *col, *prevCol;

    prevCol = _headColumn - _numColumns;
    if (prevCol < _column)
        prevCol += MAX_WINDOW_WIDTH;

    prevCol->_bestScore = 1.0;
    col = prevCol->_nextCol;
    assert(col->_numCells == 1);
    for (int ix = 1; ix < _numColumns;
            ++ix, prevCol = col, col = col->_nextCol)
    {
        for (int jx = 0; jx < col->_numCells;
                ++jx, prevCol = prevCol->_prevCol)
        {
            SEQUENCE *seq = col->_cell[jx]._seq;
            float Pword = seq->_accumScores / seq->_inCount;
            float score = prevCol->_bestScore;

            score *= pow(Pword, jx + 1);
            if (!jx || col->_bestScore < score)
            {
                col->_bestScore = score;
                col->_bestLength = jx + 1;
            }
        }
    }
}
```

```
//
// Detach one or more words from the left side of the event window
// and send these words to the output stream.
//
void DetachWords()
{
    COLUMN *col;
    SEQUENCE *stack[MAX_WINDOW_WIDTH];
    SEQUENCE **seq = &stack[ARRAY_LENGTH(stack)];
    int len, offset = 1;

    // Determine which words to detach from event window.
    *--seq = NULL;
    while (offset < _numColumns)
    {
        col = _headColumn - offset;
        if (col < _column)
            col += MAX_WINDOW_WIDTH;

        len = col->_bestLength;
        if (offset > _minColumns / 2)
            *--seq = col->_cell[len-1]._seq;

        offset += len;
    }
    if (!*seq)
    {
        if (_numColumns != MAX_WINDOW_WIDTH)
            return;

        printf("-");  // Indicate character is forced out.
        *--seq = _headColumn->_nextCol->_cell[0]._seq;
    }

    // Send detached words to output stream.
    while (*seq)
    {
        UINT8 buffer[MAX_WINDOW_WIDTH];
        UINT8 *str = &buffer[ARRAY_LENGTH(buffer)];

        *--str = '\0';
        _numColumns -= (*seq)->_info._length;
        for (SEQUENCE *p = *seq++; p; p = p->_prevSeq)
            *--str = p->_info._event;

        printf("%s%c", str, *seq ? ' ' : '\n');
    }
    col = _headColumn - _numColumns + 1;
    if (col < _column)
        col += MAX_WINDOW_WIDTH;

    // Align left edge of event window to new word boundary.
    for (len = 1;
         len <= _numColumns && col->_numCells > len;
         ++len, col = col->_nextCol)
    {
        col->_numCells = len;
        if (col->_bestLength > len)
            col->_bestLength = len;
    }
}

//
```

```c
// A new character just arrived in the input stream. Process this event.
//
void ProcessEvent(UINT8 event)
{
    COLUMN *col;
    SEQUENCE *seq;

    if (!_headColumn)
    {
        _numColumns = 1;
        _headColumn = &_column[0];
        _headColumn->_numCells = PopulateNewColumn(event, NULL,
                                                   0, _headColumn->_cell);
        return;
    }

    col = _headColumn;
    _headColumn = _headColumn->_nextCol;
    memset(_headColumn->_cell, 0, sizeof(_headColumn->_cell));
    _headColumn->_event = event;
    _headColumn->_numCells = PopulateNewColumn(event, col->_cell,
                                               col->_numCells,
                                               _headColumn->_cell);
    assert(_headColumn->_numCells < MAX_WINDOW_WIDTH);
    SetFirstLevelScores();
    if (!_enableWrite)
        return;

    if (++_numColumns <= _minColumns)
        return;

    if (_minColumns)
    {
        if (_numColumns != MAX_WINDOW_WIDTH)
        {
            seq = col->_cell[col->_bestLength - 1]._seq;
            if (seq->_accumScores / seq->_inCount < 0.50)
                return;
        }
        else
            printf("+");  // Indicate event window overflow.
    }
    SetSecondLevelScores();
    DetachWords();
}

//
// Main program: Send stream of test data to word-segmentation algorithm.
//
int main()
{
    int ix, charCount = 0;
    char *s;

    srand(123456);
    Initialize();

    // Learning phase
    for (ix = 0; ix < NUM_LEARNING_WORDS; ++ix)
    {
        int index = rand() % ARRAY_LENGTH(_testData);

        for (s = _testData[index]; *s != '\0'; ++s)
```

```
                ProcessEvent(*s);
        }

        _enableWrite = 1;  // Enable output stream.
        _headColumn = NULL;  // Reset event window.

        // Output phase
        for (ix = 0; ix < ARRAY_LENGTH(_testData); ++ix)
        {
                for (s = _testData[ix]; *s != '\0'; ++s)
                {
                        ProcessEvent(*s);
                        ++charCount;
                }
        }

        // Flush remaining words in event window to output stream.
        _minColumns = 0;
        ProcessEvent('x');

        printf("\nSEQUENCES STORED = %d", _allocCount);
        printf("\nSEQUENCES FIRING = %d", _fireCount);
        printf("\nTOTAL EVENT COUNT = %d", _eventCount);
        printf("\nSAMPLE LENGTH = %d words, %d chars\n",
                ARRAY_LENGTH(_testData), charCount);
        return 1;
}
```

The **GetHashIndex** function in the preceding program uses the ELF hash algorithm, as described by Binstock [1].

## *Acknowledgment*

A careful reading by Dana M. Van Aken of an early manuscript resulted in a number of improvements in the presentation of the concepts in this paper. The work described in this paper was inspired by the description of sequence memory by Hawkins & Blakeslee [9].